\documentclass[11pt]{article}
\usepackage[a4paper]{geometry}
\usepackage{clic2021} 
\usepackage{times} 
\usepackage{xurl} 
\usepackage[greek,english]{babel}
\usepackage{latexsym}
\usepackage{xcolor}
\usepackage{tabularx}
\usepackage{graphicx}
\usepackage{bm}
\usepackage[T3,OT2,T1]{fontenc}
\usepackage{enumitem}
\usepackage{hyperref}

\newcommand\aprilLong{April Fools' Day}
\newcommand\april{AFD}

\makeatletter
\newcommand{\smallsym}[2]{#1{\mathpalette\make@small@sym{#2}}}
\newcommand{\make@small@sym}[2]{%
  \vcenter{\hbox{$\m@th\downgrade@style#1#2$}}%
}
\newcommand{\downgrade@style}[1]{%
  \ifx#1\displaystyle\scriptstyle\else
    \ifx#1\textstyle\scriptstyle\else
      \scriptscriptstyle
  \fi\fi
}
\makeatother

\makeatletter
\def\footnoterule{\kern-3\p@
  \hrule \@width 2in \kern 2.6\p@} 
\makeatother
\newcommand{\copyrightnotice}[1]{{%
  \renewcommand{\thefootnote}{}
  \footnotetext[0]{#1}%
}}

\title{Linguistic Cues of Deception in a Multilingual \aprilLong\ Context}

\author{\textbf{Katerina Papantoniou$^{1,2}$, Panagiotis Papadakos$^{2}$, Giorgos Flouris$^{2}$}, Dimitris Plexousakis$^{1,2}$  \\
  1. Computer Science Department, University of Crete, Greece \\
  2. Institute of Computer Science, FORTH, Greece \\
  {\tt \{papanton, papadako, fgeo, dp\}@ics.forth.gr}}

\date{}

\begin{document}
\maketitle

\copyrightnotice{Copyright \copyright 2021 for this paper by its authors. Use permitted under Creative Commons License Attribution 4.0 International (CC BY 4.0).}

\begin{abstract}
\textbf{}~
In this work we consider the collection of deceptive April Fools' Day (AFD) news articles as a useful addition in existing datasets for deception detection tasks.
Such collections have an established ground truth
and are relatively easy to construct  across languages.
As a result, we introduce a corpus that includes diachronic \april\ and normal articles from Greek newspapers and news websites. On top of that, we build a rich linguistic feature set, and analyze  and compare its deception cues with the only \april\ collection currently available, which is in English. 
Following a current research thread, we also discuss the individualism/collectivism  dimension in deception with respect to these two datasets.
Lastly, we build classifiers by testing various monolingual and crosslingual settings. The results showcase that \april\ datasets can be helpful in deception detection studies,
and are in alignment with the observations of other deception detection works.





\end{abstract}

\section{Introduction}

\aprilLong\ (for short \april) is a long standing custom,
mostly in Western societies. It is the only day of the year when practical jokes and deception are expected. This is the case for all social interactions, including journalism, which is generally considered to aim at the presentation of truth. Every year on this day, newspapers and news websites take part in an unofficial competition to invent the most believable, but untrue story. In this respect, \april\ news articles fall into the deception spectrum, as they satisfy widely acceptable definitions of deception as in Masip et al.~\shortcite{Masip2005Deception}.

The massive participation of news media in this custom
establishes a rich corpus of deceptive articles from a diversity of sources. 
Although \april\ articles may exploit common linguistic instruments with satire news, like exaggeration, humour, irony and paralogism, they are usually considered a distinct category. This is mainly due to the fact that they also employ other mechanisms which characterize deception in general, like sophisms, and changes in cognitive load and emotions \cite{Hauch2015Meta} to deceive their audience.
\april\ articles are often believable, and there exist cases where sophisticated \april\ articles have been reproduced by major international news agencies worldwide\footnote{https://www.nationalgeographic.com/history/article/150331-april-fools-day-hoax-prank-history-holiday}.

This motivated us to extend our previous work on linguistic cues of deception 
and their relation to the cultural dimension of individualism and collectivism
\cite{papantoniou2021deception}, in the context of the \april. That work examines if differences in the usage of linguistic cues of deception (e.g., pronouns) across cultures can be identified  and attributed to the individualism/collectivism divide.

Specifically, the contributions of this work are:
 \begin{itemize}
    \itemsep0em 
    \item A new corpus that includes diachronic \april\ and normal articles from Greek newspapers and news websites\footnote{The collection is available in: \url{https://gitlab.isl.ics.forth.gr/papanton/elaprilfoolcorpus}}, adding one more \april\ collection to the currently unique one in English \cite{Dearden2019EnApril}.
    \item A study and discussion of the linguistic cues of deception that prevail in the Greek and English collection, along with their similarities.
    \item A discussion on whether the consideration of the individualism/collectivism cultural dimension in the context of \april\ aligns with the results of our previous work.
    \item An examination of the performance of various classifiers in identifying  \april\ articles, including  multilanguage setups.
 \end{itemize}

\section{Related work}

The creation of reliable and realistic ground truth datasets for the deception detection task is a challenging task \cite{Fitzpatrick2012BuildingAD}. 
Crowdsourcing, in the form of online campaigns in which people express themselves in truthful and/or deceitful manner for a small payment are a well established way to collect deceptive data \cite{Ott2011OpSpam}. Real-life situations such as trials \cite{Soldner2019Box} or the use of data from board games have also been employed \cite{Peskov2020Diplomacy}. Also a  popular approach is the reuse of content from sites that debunk articles like fake news and hoaxes \cite{Wang2017LIAR,Kochkina2018PHEME}. Lastly, satire news are another way to collect deceptive texts, but with some particularities due to humorous deception \cite{Skalicky2020Bluff}. 

The only work that explores \april\  articles is that of Dearden et al.~\shortcite{Dearden2019EnApril}. They collected 519 \april\ and 519 truthful stories and articles in English for a period of 14 years. A large set of features was exploited to identify deception cues in \april\  stories. Structural complexity and level of detail were among the most valuable features while the exploitation of the same feature set to a fake news dataset resulted in similar observations.

To the best of our knowledge, the only deception related dataset for the Greek language is that of Karidi et al.~\shortcite{Karidi2019Phony}. This work proposed an automatic process for the creation of a fake news and hoaxes articles corpus, but unfortunately the created corpus over Greek websites is not available. If we also consider that the creation of a Greek dataset for deception through crowdsourcing is a cumbersome and expensive task, that is further hindered by the exceptionally limited number of native Greek crowd workers, it is easy to understand why there is a lack of datasets.

Regarding the individualism/collectivism cultural dimension,
it constitutes  a well-known division of cultures that concerns the degree 
in which members of a culture value more individual over group goals and vice versa. 
In individualism, ties between individuals are loose and individuals are expected to 
take care of only themselves and their immediate families, whereas in collectivism ties in society are stronger. 
In Papantoniou et al.~\shortcite{papantoniou2021deception} there is an preliminary effort driven by prior work in psychology discipline \cite{Taylor2017CultureMC} 
to examine if deception cues are altered across cultures and if this can be attributed to this divide. 
Among the conclusions were that people from individualistic cultures employ 
more third and less first person pronouns to distance themselves from the deceit when they are deceptive, 
whereas in the collectivism group this trend is milder, signalling the effort of the deceiver to distance the group from the deceit. 
In addition, in individualistic cultures positive sentiment is employed in deceptive language,   
whereas in collectivists there is a restraint of expression of sentiment both in truthful and deceptive texts.

To this end, this work explores the deception-related characteristics of a new Greek  corpus based on \april\ articles from a variety of sources, 
and compares them with the  English ones\footnote{We also experimented with data from the limited number of satirical and hoaxes sources of the Greek Web. 
We do not discuss them here though, since the classifiers reported excellent accuracy
showcasing the lack of diversity and the existence of domain specific information in the collected data.}. 
Further, since related studies \cite{triandis_1972_interpersonal,geerthofstede1980,Koutsantoni2005Culture} describe Greece as a culture with more collectivistic characteristics 
(by using country as proxy from culture), we also discuss differences in deception cues along this cultural dimension.

\section{Corpus creation}

The \april\ articles have been hand gathered because a crawling based  collection approach was not applicable in our case. Since the news web sites industry in Greece is not huge to establish an acceptable number of crawled \april\ articles, 
we had to additionally collect articles from the press, including articles from the pre-WWW era.
Specifically, we visited the local library that maintains a printed archive of newspapers and searched for disclosure articles in the issues after the 1st April, took photos of the \april\  articles, and then used OCR and manual inspection to extract the text. 
In addition we contacted national and local news media providers to get access in their digitalized archives. 
The rest were gathered from the Web.

The articles were categorized thematically into the following five  categories: society, culture, politics, world, and sports. 
If no category was provided by the original source, we manually annotated the articles.
For each article we kept the title, the main body, the published date, the name, the type of the source (newspaper or news website), and (if available) the caption, the subtitle and the author. As preprocesing steps we applied spellcheck and normalization. 
The correction of spelling mistakes was necessary primarily for articles extracted  through OCR tools, although spelling errors were identified in other articles too. 
Normalization was performed for  homogeneity reasons in the texts retrieved from the 80's,
since we observed language differences in some forms (e.g., in the suffix of genitive case),
which are remains of an old form of Modern Greek\footnote{https://en.wikipedia.org/wiki/Katharevousa}. 

For the truthful collection we used the same manual procedure and we tried to  have a balanced dataset in terms of thematic categories.
The truthful collection consists of articles that have been published in days relatively close to the 1st of April in order to have articles that do not differ significantly in respect to their topics, mentioned named entities,  etc.

Since the \april\  tradition is vivid in Greece, we were able to locate 
a lot of such articles from various newspapers and new websites for our corpus (112 different sources).
Specifically, we managed to collect 254 truthful and 254 deceptive articles spanning over the period 1979 - 2021. In Tables~\ref{tab:statistics} to~\ref{tab:statistics_topic}  some statistics of the corpus are depicted.
\begin{table}[h] 
\caption{Overview of the dataset.}  
\centering        
\begin{tabular}{l l l }    
\hline                         
 \textbf{Measure} &\textbf{Truthful} & \textbf{Deceptive} \\ [0.5ex]    \hline                  
 Num. of articles & 254 & 254 \\
 Avg. length & 336 & 255 \\
 Min. length & 57 & 33 \\
 Max. length & 1347 & 1163 \\
\hline                          
\end{tabular} 
\label{tab:statistics} 
\end{table}

\begin{table}[h] 
\caption{Distribution of articles per topic.}  
\centering        
\begin{tabular}{l l l }    
\hline                         
 \textbf{Topic} &\textbf{Truthful} & \textbf{Deceptive} \\ [0.5ex]    \hline                  
culture & 20 & 24 \\
politics & 85 & 78 \\
society & 86 & 118 \\
sports & 22 & 29 \\
world & 41 & 5 \\
\hline                          
\end{tabular} 
\label{tab:statistics_topic} 
\end{table}

\section{Features analysis}\label{features}
For the analysis of \april\  articles we adapt and build upon the feature set used in Papantoniou et al.~\shortcite{papantoniou2021deception}, but for the Greek language. The resulting feature set consists of 64 features for the Greek language and 75 for the English, due to the smaller availability of linguistic resources for Greek (e.g., in sentiment lexicons). For the analysis we performed the non-parametric Mann–Whitney U test (two-tailed) with
a 99\% confidence interval (CI) and 
$\alpha  =  0.01$. Table~\ref{tab:stat_sign_features_greek} depicts the results of this analysis for el\april\ and en\april\ datasets\footnote{All the features are described in\\ https://gitlab.isl.ics.forth.gr/papanton/elaprilfoolcorpus}. 

In both datasets, positive sentiment is related to the deceptive articles, while negative sentiment  with the truthful articles. The only exception concerns the en\april\ dataset, where for the NRC lexicon the opposite holds (NRC is one of the six sentiment lexicons used for features in English).
In addition, negative emotions like anger, fear and sadness are related to truthful news articles in both datasets. The use of positive emotive language during deception may be a strategy for deceivers to maintain social harmony as noticed also by other studies \cite{newman,perez-rosas-etal-2018-automatic}. 
The difference in the use of emotional language between truthful and deceptive news is more intense in the en\april\ dataset, where five out of the eight emotions in the NRC lexicon are found statistical significant. 
This is in alignment with the results in Papantoniou et al.~\shortcite{papantoniou2021deception} for individualistic and collectivistic cultures. 

Further, deceptive texts seem to be related with an increased use of adverbs in both datasets. This can be related to the less concreteness of deceptive texts as discussed in Kleinberg et al.~\shortcite{kleinberg2019NE} and it is in line  with many theories of deception like the Reality Monitoring \cite{Johnson1998RealityMonitoring}, Criteria based Content Analysis \cite{Undeutsch1989} and Verifiability Approach \cite{Nahari2014Verifiability}. This also explains the prevalence of the number of named entities, spatial related words, conjunctions and  WDAL imagery score in truthful texts in the en\april\  dataset and the use of more motion verbs in deceptive texts in the el\april\ dataset. According to cognitive load theory \cite{Sweller2011CognitiveTheory} in deceptive texts the language is less specific and consists of  simpler constructs. The same holds for modality, another common feature among the datasets, that is considered a signal of subjectivity that provides a degree of uncertainty. In addition, hedges in en\april\ dataset, also express some feeling of doubt or hesitancy.

Lexical diversity as expressed by the token-type ratio (TTR), that is the ratio of unique words to the total number of tokens, is related to the deceptive texts.  This seems to contradict all the above, but could be attributed to the fact that deceptive texts are shorter. Although this is more evident in the case of the en\april\ dataset, it also holds for el\april\ dataset (see Table~\ref{tab:statistics}).

Boosters, which are words that express confidence (e.g., certainly) are quite discriminative for deceptive texts for the en\april\ dataset. Moreover we observe the connection of the future tense with deception and of the past with truth.
The above were also marked in Papantoniou et al.~\shortcite{papantoniou2021deception} in different domain from the news articles domain. 

Finally, first personal pronouns have been found to be rather discriminative of deceptive texts
in various deception detection and cultural studies,
including Papantoniou et al.~\shortcite{papantoniou2021deception}.
However, in this study pronouns are statistical important only for the en\april\ dataset. 
This probably reflects idiosyncrasies of the news domain, since articles mainly present objectively facts  
and not  opinions, and as a result the use of first personal pronouns is avoided. 
This holds for the el\april\ dataset that includes \april\ articles from the news sites and the press,
and not for the en\april\ dataset that consists of various types of \april\ articles and stories 
collected from the web through crowdsourcing\footnote{https://aprilfoolsdayontheweb.com/2004.html}.


\begin{table}[t!]
\vspace{-2mm}
\caption{The statistical significant features (p<0.1) with at least a small effect size (r>0.1) for the el\april\ and en\april\ datasets. The features are in ascending p value order. We also report the effect size. Features with moderate effect size (r>0.3)  are bold, while common features between the datasets are underlined. pp denotes personal pronouns.}  
\centering        
\resizebox{0.5\textwidth}{!}{%
\begin{tabular}{ l l }    
   \hline                      
 Deceptive & Truthful \\ [0.5ex]    \hline

 \multicolumn{2}{c}{\textit{el\april}\ }\\  \hline 
\textbf{\underline{adverbs}} (0.31)& punctuation (-0.17)\\
\underline{adj. \& adv.} (0.27) & \underline{nrc sadness}(-0.17) \\
\underline{TTR} (0.27) & plosives (-0.16) \\
\underline{pos. sentiment} (0.21)& \underline{nrc anger} (-0.15)\\
\underline{modal verbs} (0.17)& \underline{nrc fear} (-0.14)\\
motion verbs (0.117)&vowels (-0.14) \\
&\underline{consonants} (-0.14) \\  [0.5ex]     \hline

 \multicolumn{2}{c}{\textit{en\april\ }}\\ \hline

\textbf{boosters} (0.39) & NE num. (-0.27)\\
\underline{\textbf{modal verbs}} (0.35) & spatial num. (-0.26)\\
\underline{\textbf{TTR}}(0.31) & conjuctions (-0.24) \\
future (0.27) & \underline{nrc fear} (-0.23)\\
\underline{adverbs} 	(0.2) & past (-0.23)\\
1st pers. pp  (0.2)& \underline{nrc sadness} (-0.23)\\
\underline{mpqa pos.} (0.2)& \underline{nrc anger} (-0.21) \\
\underline{nrc neg.*} (-0.2) & nrc trust (-0.21)\\
2nd pers. pp (0.19) & avg. word len. (-0.17) \\
1st pers. pp pl. (0.18) &  collectivism (-0.16) \\
\underline{sentiwordnet pos.} (0.17) & \underline{nrc pos.*} (-0.16) \\
demonstrative (0.17) & wdal imagery (-0.15) \\
hedges (0.17) & mpqa neg. -0.14) \\
\underline{adj \& adv} (0.16) & nasals (-0.14) \\
present  (0.15)& fbs neg. (-0.14) \\
\underline{vader sentiment} (0.14) & \underline{consonants} (-0.13)\\
verb num. (0.14) & anew arousal (-0.13) \\
pers. pron. (0.12)& prepositions (-0.12) \\
total pronouns (0.11) & fricatives (-0.11) \\
 & 3rd per. pp sg. (-0.11) \\
  & avg. preverb len. (-0.11) \\                  
  &   nrc disgust (-0.1) \\                           
\hline                          
\end{tabular} 
\label{tab:stat_sign_features_greek} 
}
\vspace{-5mm}
\end{table}

\section{Classification}
We evaluated the predictive performance of different feature sets and approaches for \april\ datasets, including logistic regression experiments\footnote{We employ the Weka API \cite{hall09:_weka_data_minin_softw}} and  fine-tuned monolingual BERT models for each language\footnote{We used tensorflow 2.2.0, keras 2.3.1, and the bert-for-tf2 0.14.4 implementation of google-research/bert, over an AMD Radeon VII card and the ROCm 3.7 platform.} \cite{devlin-etal-2019-bert,Koutsikakis2020bert}. We also performed cross lingual experiments by exploiting the multilingual BERT model (mBERT) to examine if there are  similarities among \april\ datasets captured by the BERT. 

A stratified split to the datasets was used to create training, testing, and validation subsets with a 70-20-10 ratio. For the cross lingual experiment we trained and validated a model over the 80\% and 20\% of a language specific dataset respectively, and then tested the performance of the model over the other dataset. We report  the  results  on  test  sets,  while  validation  subsets  were  used  for  fine-tuning  the  hyper-parameters of the algorithms. For the logistic regression the tuned through brute force parameters  were: a) Weka algorithm ($SimpLog|Log$: \textit{simple logistic} \cite{Landwehr2005} or \textit{logistic} \cite{leCessie1992}) b) all n-grams of size in $[a,b]$, with $a \geq b$ and $a$, $b$ $\in$ $[1,3]$ ($(a,b)$), c) stemming ($stem$), d) attribute selection ($attrsel$) (applicable only to $Log$ algorithm since it is the default for $SimpLog$ ), e) stopwords removal ($stop$) and, f) lowercase conversion ($lowercase$).
For the BERT experiments, the hyperparameters were tuned by random sampling 60 combinations of values, keeping the combination that gave the minimum validation loss. Early stopping with patience 4 was used and the max epochs number was set to 20. The tuned hyperparameters were: learning rate, batch size, dropout rate, max token length, and randomness seeds.

In all cases, we report Recall ($R$), Precision ($P$), F-measure ($F$), Accuracy ($A$) and AUC ($A'$). 
Since the datasets are balanced the majority baseline is 50\%. The input for the models consists of the concatenation of the title, the subtitle, 
the  body of the articles and  the caption text. 
Since titles are important for deception detection \cite{Horne2017Title}
and BERT processes texts of up to 512 wordpieces,
we placed the title first.

\subsection{Logistic regression experiments}
The examined features sets were: a) the features presented in section~\ref{features} (ling), b) n-grams
features i.e., phoneme-gram (ph-gram), character-gram (char-gram), word-gram (w-gram), POS-gram (pos-gram), and syntactic-gram (sn-gram) (the latter for the  en\april\ only), and c) the linguistic+ model that represents the best model that combines the linguistic features
with any of the n-gram features. The results are presented in Tables~\ref{elAprilFoolresults} and~\ref{enAprilFoolresults}. With * we mark the setups with a statistically significant difference to the best setup regarding accuracy, based on a  two proposition z-test (1-tailed) with a 99\% CI.
We observe that the combination of \textit{lingustic} features with uni/bi/tri-grams for the el\april\ dataset and the unigrams for the en\april\ are the best setups. For the en\april\ dataset, the second best model is the combination of \textit{linguistic} features with trigrams. 
$SimpLog$ seems to perform better, while stemming, lowercase conversion and stopwords removal are generally beneficiary.

\begin{table}[t!]
\vspace{-2mm}
\caption{Logistic regression results for el \april\ .}  
\centering        
\resizebox{0.5\textwidth}{!}{%
\begin{tabular}{llllll}    
\hline                         
\textbf{Best setup}&\textbf{R}&\textbf{P}&\textbf{F}&\textbf{A'}&\textbf{A}\\  [0.5ex]    \hline               ling.$_{SimpLog}$ 	   	            &	62 	&	76	&	68	&	82	&	71	\\
ph-gram$_{(1,2),attrsel,Log}$*		 	&	70	&	67	    &	68	&	77	&	68	\\
char-gram$_{(3,3),SimpLog}$*	 	        &	72	&	68	&	70	&	76	&	69	\\
w-gram$_{(1,2),SimpLog}$	   	 &	68 &	73	&71 	& 80	&72	\\
pos-gram$_{(2,3),SimpLog}$* 	   		        &	72	&	65	&	68 	&	75	&	67	\\
ling.+$_{word,(1,3),stop,}$\\$_{lowercase,SimpLog}$ 	   		    &	\textbf{74} 	&	\textbf{79}	&	\textbf{76} 	&	 	\textbf{85}	&	\textbf{77}	\\  
\hline                          
\end{tabular} 
\label{elAprilFoolresults} 
\vspace{-3mm}
}
\end{table} 

\begin{table}[t!] 
\vspace{-2mm}
\caption{Logistic regression results for en \april\ .}  
\centering        
\resizebox{0.5\textwidth}{!}{%
\begin{tabular}{llllll}    
\hline                         
\textbf{Best setup}&\textbf{R}&\textbf{P}&\textbf{F}&\textbf{A'}&\textbf{A}\\  [0.5ex]    \hline               ling.$_{Log}$* 	   	            &	66 	&	80	&	72	&	\textbf{87}	&	75	\\
ph-gram$_{(1,1),SimpLog}$	 	&	80	&	77	    &	78	&	84	&	78	\\
char-gram$_{(1,3),attrsel,Log}$* 	&		       76	&	72	&	74	&	80	&	73	\\
w-gram$_{(1,1),stem,SimpLog}$	    &	\textbf{79} &	\textbf{81}	&\textbf{80}	& \textbf{87}	&\textbf{80}	\\
pos-gram$_{(3,3),SimpLog}$*		            &	71	&	69	&	70	&	76	&	69	\\
sn-gram$_{(2,2),SimpLog}$* 	 	           &	80	&	68	&	73 	&	77	&	71	\\
ling.+$_{Word,(3,3),stop,}$
\\$_{lowercase,SimpLog}$	    & 	    	74 	&	80	&	77 	&	 	\textbf{87}	&	78	\\  
\hline                          
\end{tabular} 
\label{enAprilFoolresults} 
}
\vspace{-5mm}
\end{table} 

\begin{table}[ht!] 
\vspace{-2mm}
  \caption{BERT models evaluation results. }  
\centering        
\begin{tabular}{llllll}    
\hline                         
&\textbf{R}&\textbf{P}&\textbf{F}&\textbf{A'}&\textbf{A}\\  [0.5ex]    \hline    
el$_{bert}$&76	&83 &79	&90
&\textbf{80} \\
el$_{bert+ling}$&72	&80 &76	&86 & 77 \\
el$_{mbert}$&70	&73 &71	&81&72\\ 
el$_{mbert+ling}$&50	&81 &62 &83&69\\ 
en$_{bert}$&88	&85 &87	&94
&\textbf{86} \\
en$_{bert+ling}$&74	&89 &81	&91 & 83 \\
en$_{mbert}$&50	&95 &66	&91&73 \\
en$_{mbert+ling}$&50	&86 &63	&86&71\\
en$\rightarrow$el $_{mbert}$&38	&76 &50	&72 & 63 \\
el$\rightarrow$en $_{mbert}$&24	&87 &37	&71 & 60 \\
\hline                          
\end{tabular} 
\label{berT} 
\vspace{-5mm}
\end{table} 

\subsection{BERT experiments}
In these experiments, we fine-tuned BERT by adding a task-specific linear classification layer on top,  using the sigmoid activation function. We also combined BERT with linguistics features by  concatenating the embedding of the [CLS] token with the linguistic features, and pass the resulting vector to the task-specific classifier (with a slightly modified architecture). The results of the experiments are presented in Table~\ref{berT}. Although it outperformed logistic regression experiments in both datasets, 
the differences are not statistical significant. In addition, the combination with linguistic features is not beneficial.
Multilingual BERT models perform worse, especially for Greek.
In the cross lingual experiments the classifiers performance is limited to about 60\% accuracy in both experiments, showcasing that the BERT layers are not able to capture language agnostic information from our datasets. 

\section{Conclusion and Future work}
We introduced a new dataset with \april\ news articles in Greek and
analyzed and compared its deception  cues  with  another English one.
The results showcased the use of emotional language, especially of positive sentiment, 
for deceptive articles which is even more prevalent in the individualistic English dataset.
Further, deceptive articles use less concrete language, as manifested by the increased use of adverbs, hedges, and boosters 
and less usage of named entities, spatial related words and conjunctions compared to the truthful ones.
The future and past tenses were correlated with deceptive and truthful articles respectively. 
All the above, mainly align with previous work \cite{papantoniou2021deception}, except from some differences in the usage of pronouns for the Greek  dataset, 
which is attributed to the idiosyncrasies of the news domain.
The accuracy of the deployed classifiers offered adequate performance,
with no statistically significant differences between the best logistic regression and the BERT models. 

In the future we aim at creating even more crosslingual datasets for deception detection tasks through crowdsourcing and by employing the Chattack platform \cite{smyrnakis2021chattack}.

\section*{Acknowledgement}
This work has received funding from the Hellenic Foundation for Research and Innovation (HFRI) and the General Secretariat for Research and Technology (GSRT), under grant agreement No 4195.

\bibliographystyle{acl}
\bibliography{clic_papantoniou.bib}

\end{document}